\documentclass[journal]{IEEEtran}
\usepackage{times}
\usepackage{epsfig}
\usepackage{graphicx}
\usepackage{amsmath}
\usepackage{amssymb}
\usepackage{epstopdf}
\usepackage{subfigure}
\usepackage{multirow}
\usepackage{tabulary}
\usepackage{tabularx}
\usepackage{hhline}

\ifCLASSINFOpdf
\else
\fi

\begin{document}
%
\title{FoodNet: Recognizing Foods Using Ensemble of Deep Networks}
%
%
%

\author{Paritosh Pandey$^*$, Akella Deepthi$^*$, Bappaditya Mandal
        and N. B. Puhan 

\thanks{P. Pandey, A. Deepthi and N. B. Puhan are with the School of Electrical Science, Indian Institute of Technology (IIT), Bhubaneswar, Odisha 751013, India. E-mail: \{pp20, da10, nbpuhan\}@iitbbs.ac.in}
\thanks{B. Mandal is with the Kingston University, London, Surrey KT1 2EE, United Kingdom. Email: b.mandal@kingston.ac.uk}
\thanks{$^*$ Represents equal contribution from the authors.}
}
%
%

{}
%

\maketitle

\begin{abstract}
In this work we propose a methodology for an automatic food classification system which recognizes the contents of the meal from the images of the food. We developed a multi-layered deep convolutional neural network (CNN) architecture that takes advantages of the features from other deep networks and improves the efficiency. Numerous classical handcrafted features and approaches are explored, among which CNNs are chosen as the best performing features. Networks are trained and fine-tuned using preprocessed images and the filter outputs are fused to achieve higher accuracy. Experimental results on the largest real-world food recognition database ETH Food-101 and newly contributed Indian food image database demonstrate the effectiveness of the proposed methodology as compared to many other benchmark deep learned CNN frameworks.
\end{abstract}

\begin{IEEEkeywords}
Deep CNN, Food Recognition, Ensemble of Networks, Indian Food Database.
\end{IEEEkeywords}

%
\IEEEpeerreviewmaketitle

\section{Introduction and Current Approaches}
\IEEEPARstart{T}{here} has been a clear cut increase in the health consciousness of the global urban community in the previous few decades. Given the rising number of cases of health problems attributed to obesity and diabetes reported every year, people (including elderly, blind or semi-blind or dementia patients) are forced to record, recognize and estimate calories in their meals. Also, in the emerging social networking photo sharing, food constitutes a major portion of these images. Consequently, there is a rise in the market potential for such fitness apps products which cater to the demand of logging and tracking the amount of calories consumed, such as \cite{MealSnap,Eatly}. Food items generally tend to show intra-class variation depending upon the method of preparation, which in turn is highly dependent on the local flavors as well as the ingredients used. This causes large variations in terms of shape, size, texture, and color. Food items also do not exhibit any distinctive spatial layout. Variable lighting conditions and the point of view also lead to intra-class variations, thus making the classification problem even more difficult \cite{Austin1,Mandal8,Mandal12}. Hence food recognition is a challenging task, one that needs addressing.

In the existing literature, numerous methodologies assume that the texture, color and shape of food items are well defined \cite{Sasano1,Pham1,Mandal10,Martinel1}. This may not be true because of the local variations in the method of food preparation, as well as the ingredients used. Feature descriptors like histogram of gradient, color correlogram, bag of scale-invariant feature transform, local binary pattern, spatial pyramidal pooling, speeded up robust features (SURF), etc, have been applied with some success on small datasets \cite{Martinel1}. Hoashi \emph{et al.} in \cite{Hoashi1}, and Joutou \emph{et al.} in \cite{Joutou1} propose multiple kernel learning methods to combine various feature descriptors. The features extracted have generally been used to train an SVM \cite{Cortes1}, with a combination of these features being used to boost the accuracy.

A rough estimation of the region in which targeted food item is present would help to raise the accuracy for cases with non-uniform background, presence of other objects and multiple food items \cite{Liu13}. Two such approaches use standard segmentation and object detection methods \cite{Bola1} or asking the user to input a bounding box providing this information \cite{Kawano1}. Kawano \emph{et al.} \cite{Kawano1,Yanai1} proposed a semi-automated approach for bounding box formation around the image and developed a real-time recognition system. It is tedious, unmanageable and does not cater to the need of full automation. Automatic recognition of dishes would not only help users effortlessly organize their extensive photo collections but would also help online photo repositories make their content more accessible. Lukas \emph{et al.} in \cite{Bossard1} have used a random forest to find discriminative region in an image and have shown to under perform convolutional neural network (CNN) feature based method \cite{Krizhevsky1}.

In order to improve the accuracy, Bettadapura \emph{et al.} in \cite{Bettadapura1} used geotagging to identify the restaurant and search for matching food item in its menu. Matsuda \emph{et al.} in \cite{Matsuda1} employed co-occurrence statistics to classify multiple food items in an image by eliminating improbable combinations. There has been certain progress in using ingredient level features \cite{Wang6,Chen5,baxterfood} to identify the food item. A variant of this method is the usage of pairwise statistics of local features \cite{Shulin1}. In the recent years CNN based classification has shown promise producing excellent results even on large and diverse databases with non-uniform background. Notably, deep CNN based transferred learning using fine-tuned networks is used in \cite{Sibo1,Zhang2,Park2,Ana1,Gan2} and cascaded CNN networks are used in \cite{Zhang3}. In this work, we extend the CNN based approaches towards combining multiple networks and extract robust food discriminative features that would be resilient against large variations in food shape, size, color and texture. We have prepared a new Indian food image database for this purpose, the largest to our knowledge and experimented on two large databases, which demonstrates the effectiveness of the proposed framework. We will make all the developed models and Indian food database available online to public \cite{Mandal15}. Section II describes our proposed methodology and Section III provides the experimental results before drawing conclusions in Section IV.

\section{Proposed Method}
Our proposed framework is based on recent emerging very large deep CNNs. We have selected CNNs because their ability to learn operations on visual data is extremely good and they have been employed to obtain higher and higher accuracies on challenges involving large scale image data \cite{Olga1}. We have performed extensive experiments using different handcrafted features (such as bag of words, SURF, etc) and CNN feature descriptors. Experimental results show that CNNs outperform all the other methods by a huge margin, similar to those reported in \cite{Martinel1} as shown in Table \ref{Table1}. It can be seen that CNN based methods (SELC \& CNN) features performs much better as compared to others.
\begin{table}[!htb] \tiny
      \centering
        \caption{\tiny Accuracy (\%) of handcrafted \& CNN features on ETH Food-101 database. The methods are SURF + Bag of Words 1024 (BW1024), SURF + Independent Fischer Vector 64 (SURF+IFV64), Bag of Words (BW), Independent Fischer Vectors (IFV), Mid-level Discriminative Superpixel (MDS), Random Forest Discriminative Component (RFDC), Supervised Extreme Learning Committee (SELC) and AlexNet trained from scratch (CNN).} \label{Table1}
        \begin{tabular}{|c|c|c|c|c|c|c|c|c|}
        \hline
        Methods & BW1024 & SURF+IFV64 & BW & IFV & MDS & RFDC & SELC & CNN\\
        \hline
        Top-1 & 33.47 & 44.79 & 28.51 & 38.88 & 42.63 & 50.76 & 55.89 & 56.40\\
        \hline
        \end{tabular}
\end{table}

\subsection{Proposed Ensemble Network Architecture}
We choose AlexNet architecture by Krizhevsky \emph{et al.} \cite{Krizhevsky1} as our baseline because it offers the best solution in terms of significantly lesser computational time as compared to any other state-of-the-art CNN classifier. GoogLeNet architecture by Szegedy \emph{et al.} \cite{Christian1} uses the sparsity of the data to create dense representations that give information about the image with finer details. It develops a network that would be deep enough, as it increases accuracy and yet have significantly less parameters to train. This network is an approximation of the sparse structure of a convolution network by dense components. The building blocks called Inception modules, is basically a concatenation of filter banks with a mask size of $1\times1$, $3\times3$ and $5\times5$. If the network is too deep, the inception modules lead to an unprecedented rise in the cost of computation. Therefore, $1\times1$ convolutions are used to embed the data output from the previous layers.

ResNet architecture by He \emph{et al.} \cite{He4} addresses the problem of degradation of learning in networks that are very deep. In essence a ResNet is learning on residual functions of the input rather than unreferenced functions. The idea is to reformulate the learning problem into one that is easier for the network to learn. Here the original problem of learning a function $H(x)$ gets transformed into learning non-linearly by various layers fitting the functional form $H(x) = \Gamma(x) + x$, which is easier to learn, where the layers have already learned $\Gamma(x)$ and the original input is $x$. These CNN networks are revolutionary in the sense that they were at the top of the leader board of ImageNet classification at one or other time \cite{Olga1}, with ResNet being the network with maximum accuracy at the time of writing this paper. The main idea behind employing these networks is to compare the increment in accuracies with the depth of the network and the number of parameters involved in training. Our idea is to create an ensemble of these classifiers using another CNN on the lines of a Siamese network \cite{Bromley1} and other deep network combinations \cite{Zuo1}.
\begin{figure}
\centering
\includegraphics[width=1.0\linewidth]{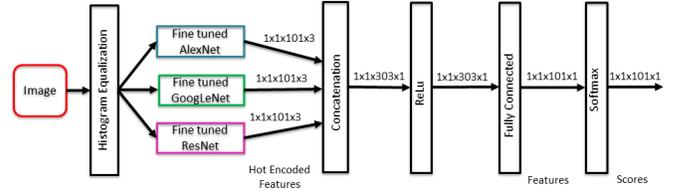}
\caption{Our proposed CNN based ensemble network architecture.} \label{Fig1:NetworkArchi}
\end{figure}

In a Siamese network \cite{Bromley1}, two or more identical subnetworks are contained within a larger network. These subnetworks have the same configuration and weights. It has been used to find comparisons or relationships between the two input objects or patches. In our architecture, we use this idea to develop a three layered structure to combine the feature outputs of three different subsections (or subnetworks) as shown in Fig. \ref{Fig1:NetworkArchi}. We hypothesize that these subnetworks with proper fine-tuning would individually contribute to extract better discriminative features from the food images. However, the parameters along with the subnetwork architectures are different and the task is not that of comparison (as in case of Siamese network \cite{Bromley1}) but pursue classification of food images. Our proposition is that the features once added with appropriate weights would give better classification accuracies.

Let $I(w,h,c)$ represents a pre-processed input image of size $w\times h$ pixels to each of the three fine-tuned networks and $c$ is the number of channels of the image. Color images are used in our case. We denote $C(m,n,q)$ as the convolutional layer, where $m$ and $n$ are the sides length of the receptive field and $q$ is the number of filter banks. Pooling layer is denoted by $P(s,r)$, where $r$ is the side length of the pooling receptive field and $s$ is the number of strides used in our CNN model. In our ensemble net we did not use pooling. But in our fine-tuned networks pooling is employed with variable parameters. GoogLeNet for example uses overlapping pooling in the inception module. All convolution layers are followed by ReLU layers (see the text in Sec \ref{NetworkDetails}) considered as an in-built activation. $L$ represents the local response normalization layer. Fully connected layer is denoted by $F(e)$, where $e$ is the number of neurons. Hence, the AlexNet CNN model after fine-tuning is represented as:

{\tiny
\begin{equation}\label{AlexNet}
\begin{aligned}
\Phi_A \equiv & ~I(227,227,3) \longrightarrow C(11,4,96) \longrightarrow L \longrightarrow P(2,3)
\longrightarrow C(5,1,256) \\
& \longrightarrow L \longrightarrow P(2,3) \longrightarrow C(3,1,384) \longrightarrow C(3,1,384) \longrightarrow C(3,1,256) \\
&  \longrightarrow P(2,3) \longrightarrow F(4096) \longrightarrow F(4096)  \longrightarrow F(e).
\end{aligned}
\end{equation}
}
AlexNet is trained in a parallel fashion, referred as a depth of 2. Details of the architecture can be found in \cite{Krizhevsky1}. For GoogLeNet we need to define the inception module as: $D(c1, cr3, c3, cr5, c5, crM)$, where $c1$, $c3$ and $c5$ represent number of filter of size $1\times1$, $3\times3$ and $5\times5$, respectively. $cr3$ and $cr5$ represent number of $1\times1$ filters used in the reduction layer prior to $3\times3$ and $5\times5$ filters, and $crM$ represents the number of $1\times1$ filters used as reduction after the built in max pool layer. Hence GoogLeNet is fine-tuned as:

{\tiny
\begin{equation}\label{GoogleNet}
\begin{aligned}
\Phi_G \equiv & ~I(224,224,3) \longrightarrow C(7,2,64) \longrightarrow P(2,3) \longrightarrow L \longrightarrow C(1,1,64)\\
      & \longrightarrow C(3,1,192) \longrightarrow L \longrightarrow P(2,3) \longrightarrow D(64,96,128,16,32,32) \\
      & \longrightarrow D(128,128,192,32,96,64) \longrightarrow P(2,3) \longrightarrow \\
      & D(192,96,208,16,48,64) \longrightarrow D(160,112,224,24,64,64) \longrightarrow \\
      & D(128,128,256,24,64,64) \longrightarrow D(112,144,288,32,64,64) \longrightarrow \\
      & D(256,160,320,32,128,128) \longrightarrow P(2,3) \longrightarrow D(256,160,320,32,\\128,128)
      & \longrightarrow D(384,192,384,48,128,128) \longrightarrow P^*(1,7) \longrightarrow F(e),
\end{aligned}
\end{equation}
}
$P^*$ refers to average pooling rather than max pooling used everywhere else. For fine-tuned ResNet, each repetitive residual unit is presented inside as $R$ and it is defined as:

{\tiny
\begin{equation}\label{ResNet}
\begin{aligned}
\Phi_R \equiv & ~I(224,224,3) \longrightarrow C(7,2,64) \longrightarrow P(2,3) \longrightarrow 3 \times R(C(1,1,64) \\
& \longrightarrow C(3,1,64) \longrightarrow C(1,1,256)) \longrightarrow R(C(1,2,128) \longrightarrow C(3,2,128) \\
& \longrightarrow C(1,2,512)) \longrightarrow 3 \times R(C(1,1,128) \longrightarrow C(3,1,128) \\
& \longrightarrow C(1,1,512)) \longrightarrow R(C(1,2,256) \longrightarrow C(3,2,256) \longrightarrow \\
& C(1,2,1024)) \longrightarrow 5\times R(C(1,1,256) \longrightarrow C(3,1,256) \longrightarrow \\
& C(1,1,1024)) \longrightarrow R(C(1,2,512) \longrightarrow C(3,2,512) \longrightarrow C(1,2,2048)) \\
& \longrightarrow 2\times R(C(1,1,512) \longrightarrow C(3,1,512) \longrightarrow C(1,1,2048)) \\
& \longrightarrow P^*(1,7) \longrightarrow F(e).
\end{aligned}
\end{equation}
}
Batch norm is used after every convolution layer in ResNet. The summations at the end of each residual unit are followed by a ReLU unit. For all cases, the length of $F(e)$ depends on the number of categories to classify. In our case, $e$ is the number of classes. Let $F_i$ denote the features from each of the fine-tuned deep CNNs given by (\ref{AlexNet})-(\ref{ResNet}), where $i\in\{A, G, R\}$. Let the concatenated features are represented by $\Omega(O,c)$, where $O$ is the output features from all networks, given by:
\begin{equation}\label{EnsembleNet}
O=concatenate(w_i F_i)\mid \forall~i,
\end{equation}
where $w_i$ is the weight given to features from each of the networks with the constraint, such that $\Sigma_i w_i = 1$. We define the developed ensemble net as the following:
\begin{equation}\label{EnsembleNet}
\begin{aligned}
\Phi_E \equiv & ~\Omega(e*\eta,c) \longrightarrow ReLU \longrightarrow F(e) \longrightarrow SoftMax,
\end{aligned}
\end{equation}
where $\eta$ is the number of fine-tuned networks. The $SoftMax$ function or the normalized exponential function is defined as:
\begin{equation}\label{SoftMax}
S(F)_j=\frac{exp^{F_j}}{\sum_{k=1}^e exp^{F_k}}, \text{for}~ j=1, 2, \ldots, e,
\end{equation}
where $exp$ is the exponential. The final class prediction $D\in\{1, 2, \ldots, e\}$ is obtained by finding the $maximum$ of the values of $S(F)_j$, given by:
\begin{equation}\label{classPredict}
D=\arg\max_j(S(F)_j), \text{for}~ j=1, 2, \ldots, e.
\end{equation}

\subsection{Network Details}
\label{NetworkDetails}
The ensemble net we designed consists of three layers as shown in Fig. \ref{Fig1:NetworkArchi}. Preprocessed food images are used to fine-tune all the three CNN networks: AlexNet, GoogLeNet and ResNet. Then the first new layer one concatenates the features obtained from the previously networks, passing it out with a rectified linear unit (ReLU) non-linear activation. The outputs are then passed to a fully connected (fc) layer that convolves the outputs to the desired length of the number of classes present. This is followed by a softmax layer which computes the scores obtained by each class for the input image.

The pre-trained models are used to extract features and train a linear kernel support vector machine (SVM). The feature outputs of the fully connected layers and max-pool layers of AlexNet and GoogLeNet are chosen as features for training and testing the classifiers. For feature extraction, the images are resized and normalized as per the requirement of the networks. For AlexNet we used the last fully connected layer to extract features (fc7) and for GoogLeNet we used last max pool layer (cls3\_pool). On the ETH Food 101 database, the top-1 accuracy obtained remained in the range of 39.6\% for AlexNet to 44.06\% for GoogLeNet, with a feature size varying from a minimum of 1000 features per image to 4096 features per image. Feature length of the features extracted out of the last layer is 1000. The feature length out of the penultimate layer of AlexNet gave a feature length of 4096 features, while the ones out of GoogLeNet had a feature length of 1024. All the three networks are fine-tuned using the ETH Food-101 database. The last layer of filters is removed from the network and replaced with an equivalent filter giving an output of the size $1\times1\times101$, i.e., a single value for 101 channels. These numbers are interpreted as scores for each of the food class in the dataset. Consequently, we see a decrease in the feature size from $1\times1000$ for each image to $1\times101$ for each image. AlexNet is trained for a total of 16 epochs.

We choose the MatConvNet \cite{Vlfeat3} implementation of GoogLeNet with maximum depth and maximum number of blocks. The implementation consists of 100 layers and 152 blocks, with 9 Inception modules (very deep!). To train GoogLeNet, the deepest softmax layer is chosen to calculate objective while the other two are removed. The training ran for a total of 20 epochs. ResNet's smallest MatConvNet model with 50 layers and 175 blocks is used. The capacity to use any deeper model is limited by the capacity of our hardware. The batch size is reduced to 32 images for the same reason. ResNet is trained with the data for 20 epochs. The accuracy obtained increased with the depth of the network. The ensemble net is trained with normalized features/outputs of the above three networks. Parametrically weights are decided for each network feature by running the experiments multiple times. A total of 30 epochs are performed.
A similar approach is followed while fine-tuning the network for Indian dataset. As the number of images is not very high, jitters are introduced in the network to make sure the network remains robust to changes. Same depth and parameters are used for the networks. The output feature has a length of $1\times1\times50$ implying a score for each of the 50 classes.

\section{Experimental Setup and Results}
The experiments are performed on a high end server with 128GB of RAM equipped with a NVDIA Quadro K4200 with 4GB of memory and 1344 CUDA cores. We performed the experiments on MATLAB 14a using the MatConvNet library offered by vlFeat \cite{Vlfeat2}. Caffe's pre-trained network models imported in MatConvNet are used. We perform experiments on two databases: ETH Food-101 Database and and our own newly contributed Indian Food Database.

\subsection{Results on ETH Food-101 Database}
ETH Food-101 \cite{Bossard1} is the largest real-world food recognition database consisting of 1000 images per food class picked randomly from foodspotting.com, comprising of 101 different classes of food. So there are 101,000 food images in total, sample images can be seen in \cite{Bossard1}. The top 101 most popular and consistently named dishes are chosen and randomly sampled 750 training images per class are extracted. Additionally, 250 test images are collected for each class, and are manually cleaned. Purposefully, the training images are not cleaned, and thus contain some amount of noise. This comes mostly in the form of intense colors and sometimes wrong labels to increase the robustness of the data. All images are rescaled to have a maximum side length of 512 pixels. In all our experiments we follow the same training and testing protocols as that in \cite{Bossard1,Martinel1}.

\begin{figure}
\centering
\includegraphics[width=1.0\linewidth]{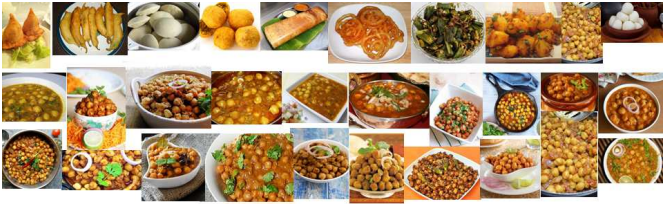}
\caption{Top row: 10 sample Indian food images. Bottom two rows: one of the food samples (1 class) variations (20 images).} \label{Fig2:FoodImages}
\end{figure}

All the real-world RGB food images are converted to HSV format and histogram equalization are applied on only the intensity channel. The result is then converted back to RGB format. This is done to ensure that the color characteristics of the image does not change because of the operation and alleviate any bias that could have been present in the data due to intensity/illumination variations.
\begin{table}[!htb] \tiny
    \begin{minipage}{.5\linewidth}
      \centering
        \caption{\tiny Accuracy (\%) for ETH Food-101 and comparison with other methods after fine-tuning.} \label{Table2}
        \begin{tabular}{|c|c|c|c|}
        \hline
        Network/Features & Top-1 & Top-5 & Top-10 \\
        \hline
        AlexNet & 42.42 & 69.46 & 80.26 \\
        \hline
        GoogLeNet & 53.96 & 80.11 & 88.04 \\
        \hline
        Lukas \emph{et al.} \cite{Bossard1} & 50.76 & - & - \\
        \hline
        Kawano \emph{et al.} \cite{Kawano1} & 53.50 & 81.60 & 89.70 \\
        \hline
        Martinel \emph{et al.} \cite{Martinel1} & 55.89 & 80.25 & 89.10 \\
        \hline
        ResNet & 67.59 & 88.76 & 93.79 \\
        \hline
        \textbf{Ensemble Net} & \textbf{72.12} & \textbf{91.61} & \textbf{95.95} \\
        \hline
        \end{tabular}
    \end{minipage}
    \begin{minipage}{.48\linewidth}
      \caption{\tiny Accuracy (\%) for Indian Food Database and comparison with other methods after fine-tuning.} \label{Table3}
      \begin{tabular}{|c|c|c|c|}
      \hline
      Network/Features & Top-1 & Top-5 & Top-10 \\
      \hline
      AlexNet & 60.40 & 90.50 & 96.20 \\
      \hline
      GoogLeNet & 70.70 & 93.40 & 97.60 \\
      \hline
      ResNet & 43.90 & 80.60 & 91.50 \\
      \hline
      \textbf{Ensemble Net} & \textbf{73.50}  & \textbf{94.40} & \textbf{97.60} \\
      \hline
      \end{tabular}
    \end{minipage}%
\end{table}

\begin{figure}
\centering
\begin{minipage}[b]{4.3cm}
\includegraphics*[height=3.4cm]{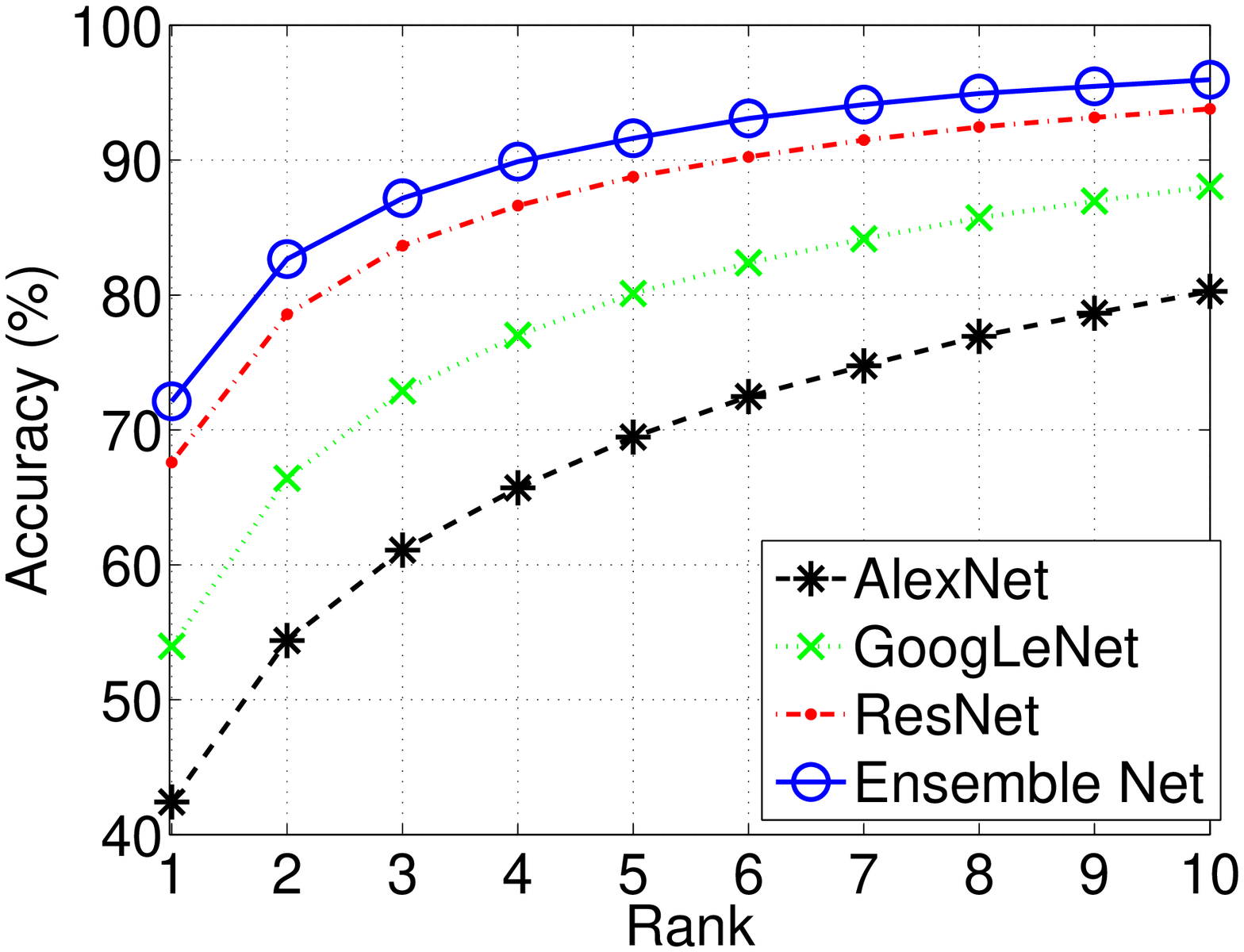}
\centering (a)
\end{minipage}
\begin{minipage}[b]{4.3cm}
\includegraphics*[height=3.4cm]{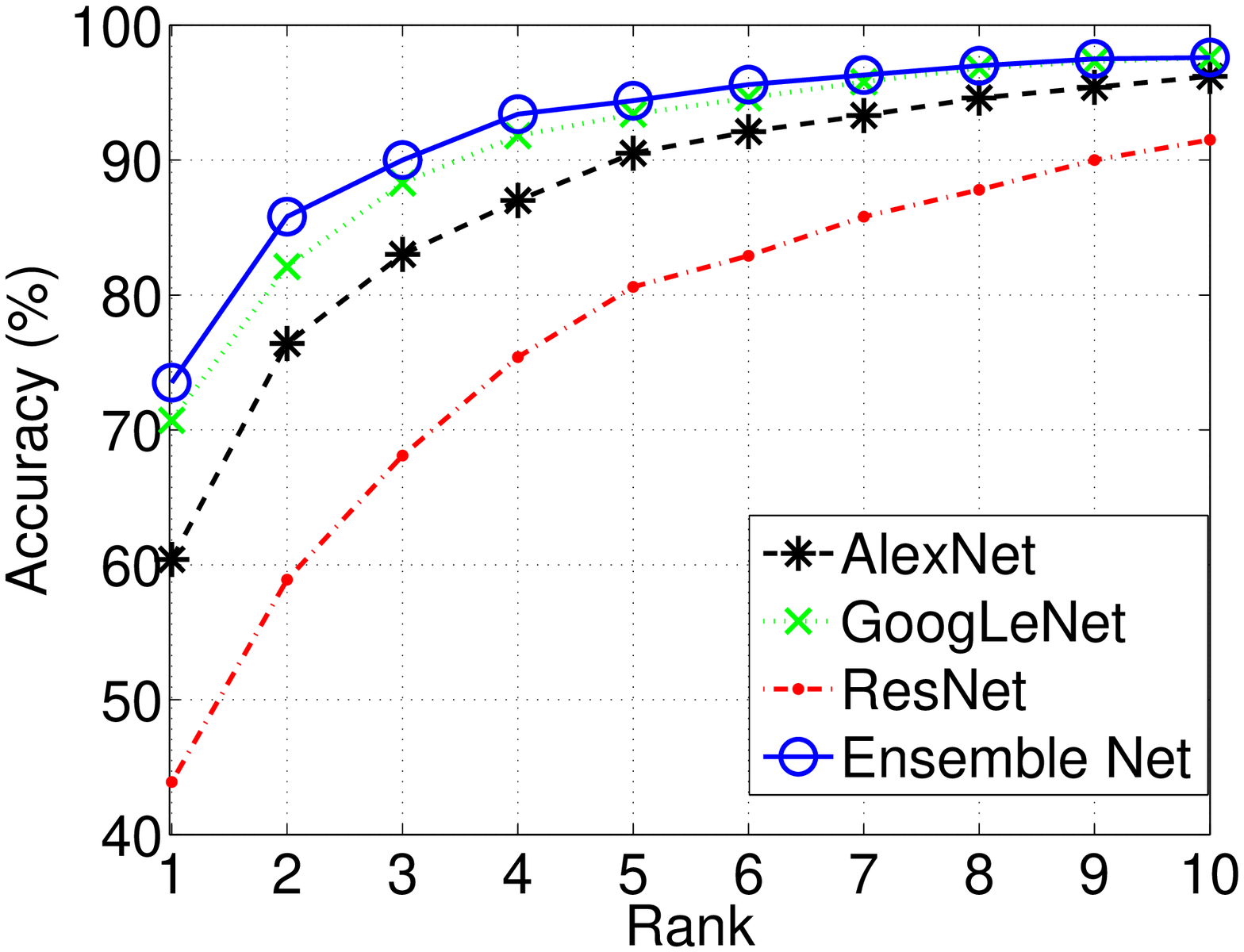}
\centering (b)
\end{minipage}
\caption{Rank vs Accuracy plots using various CNN frameworks, (a) for ETH Food 101 Database and (b) for Indian Food Database.} \label{Fig4:ETHAccuracy}
\end{figure}

Table \ref{Table2} shows the Top-1, Top-5 and Top-10 accuracies using numerous current state-of-the-art methodologies on this database. We tried to feed outputs from the three networks into the SVM classifier but the performance was not good. We have noted only the highest performers, many more results can be found in \cite{Martinel1}. It is evident that with fine-tuning the network performance has increased to a large extent. Fig. \ref{Fig4:ETHAccuracy} (a) shows accuracies with the ranks plot up to top 10, where the rank $r:r \in \{1, 2, \ldots, 10\}$ shows corresponding accuracy of retrieving at least 1 correct image among the top $r$ retrieved images. This kind of graphs show the overall performance of the system at different number of retrieved images. From Table \ref{Table2} and Fig. \ref{Fig4:ETHAccuracy} (a), it is evident that our proposed ensemble net has outperformed consistently all the current state-of-the-art methodologies on this largest real-world food database.

\subsection{Results on Indian Food Database}
One of the contributions of this paper is the setting up of an Indian food database, the first of its kind. It consists of 50 food classes having 100 images each. Some sample images are shown in Fig. \ref{Fig2:FoodImages}. The classes are selected keeping in mind the varied nature of Indian cuisine. They differ in terms of color, texture, shape and size as the Indian food lacks any kind of generalized layout. We have ensured a healthy mix of dishes from all parts of the country giving this database a true representative nature. Because of the varied nature of the classes present in the database, it offers the best option to test a protocol and classifier for its robustness and accuracy. We collected images from online sources like foodspotting.com, Google search, as well as our own captured images using hand-held mobile devices. Extreme care was taken to remove any kind of watermarking from the images. Images with textual patterns are cropped, most of the noisy images discarded and a clean dataset is prepared. We also ensured that all the images are of a minimum size. No upper bound on image size has been set. Similar to the ETH Food-101 database protocol, we have randomly selected 80 food images per class for 50 food classes in the training and remaining in the test dataset.

Fig. \ref{Fig4:ETHAccuracy} (b) shows accuracies with the ranks plot up to top 10 and Table \ref{Table3} shows the Top-1, Top-5 and Top-10 accuracies using some of the current state-of-the-art methodologies on this database. Both these depict that our proposed ensemble of the networks (Ensemble Net) is better at recognizing food images as compared to that of the individual networks. ResNet under performs as compared to GoogLeNet and AlexNet probably because of the lack of sufficient training images to train the network parameters. For overall summary: as is evident from these figures (Fig. \ref{Fig4:ETHAccuracy} (a) and (b)) and tables (Tables \ref{Table2} and \ref{Table3}) that there is no single second best method that outperforms all others methods in both the databases, however, our proposed approach (Ensemble Net) outperforms all other methods consistently for all different ranks in both the databases.


\section{Conclusions}
Food recognition is a very crucial step for calorie estimation in food images. We have proposed a multi-layered ensemble of networks that take advantages of three deep CNN fine-tined subnetworks. We have shown that these subnetworks with proper fine-tuning would individually contribute to extract better discriminative features from the food images. However, in these subnetworks the parameters are different, the subnetwork architectures and tasks are different. Our proposed ensemble architecture outputs robust discriminative features as compared to the individual networks. We have contributed a new Indian Food Database, that would be made available to public for further evaluation and enrichment. We have conducted experiments on the largest real-world food images ETH Food-101 Database and Indian Food Database. The experimental results show that our proposed ensemble net approach outperforms consistently all other current state-of-the-art methodologies for all the ranks in both the databases.

\ifCLASSOPTIONcaptionsoff
  \newpage
\fi

\bibliographystyle{IEEEtran}
\bibliography{/media/bappaditya/System/Bappaditya/Paper43/BiblioMar2017}

\end{document}